\documentclass[10pt,twocolumn,letterpaper]{article}

 \usepackage[pagenumbers]{cvpr} 

\usepackage{graphicx}
\usepackage{amsmath}
\usepackage{amssymb}
\usepackage{booktabs}
\usepackage{float}
\usepackage{multicol}
\usepackage{multirow}
\usepackage{colortbl}
\usepackage{xcolor}
\usepackage{color}
\usepackage{algorithm}
\usepackage{algpseudocode}
\usepackage[accsupp]{axessibility}
\usepackage{setspace}

\definecolor{aliceblue}{rgb}{0.94, 0.97, 1.0}

\definecolor{cvprblue}{rgb}{0.21,0.49,0.74}
\usepackage[pagebackref,breaklinks,colorlinks,citecolor=cvprblue]{hyperref}

\def\paperID{10087} 
\def\confName{CVPR}
\def\confYear{2024}

\title{A Lightweight Clustering Framework for Unsupervised Semantic Segmentation}

\author{Yau Shing Jonathan Cheung \quad Xi Chen \quad Lihe Yang \quad Hengshuang Zhao\\
The University of Hong Kong\\
}

\begin{document}
\maketitle
\begin{abstract}
Unsupervised semantic segmentation aims to categorize each pixel in an image into a corresponding class without the use of annotated data. It is a widely researched area as obtaining labeled datasets is expensive. While previous works in the field have demonstrated a gradual improvement in model accuracy, most required neural network training. This made segmentation equally expensive, especially when dealing with large-scale datasets. We thus propose a lightweight clustering framework for unsupervised semantic segmentation. We discovered that attention features of the self-supervised Vision Transformer exhibit strong foreground-background differentiability. Therefore, clustering can be employed to effectively separate foreground and background image patches. In our framework, we first perform multilevel clustering across the Dataset-level, Category-level, and Image-level, and maintain consistency throughout. Then, the binary patch-level pseudo-masks extracted are upsampled, refined and finally labeled. Furthermore, we provide a comprehensive analysis of the self-supervised Vision Transformer features and a detailed comparison between DINO and DINOv2 to justify our claims. Our framework demonstrates great promise in unsupervised semantic segmentation and achieves state-of-the-art results on PASCAL~VOC and MS~COCO datasets.
\vspace{-0.5cm}
\end{abstract}

\section{Introduction}
\label{sec:intro}
\begin{figure}[t]
    \centering
    \includegraphics[width=0.47\textwidth]{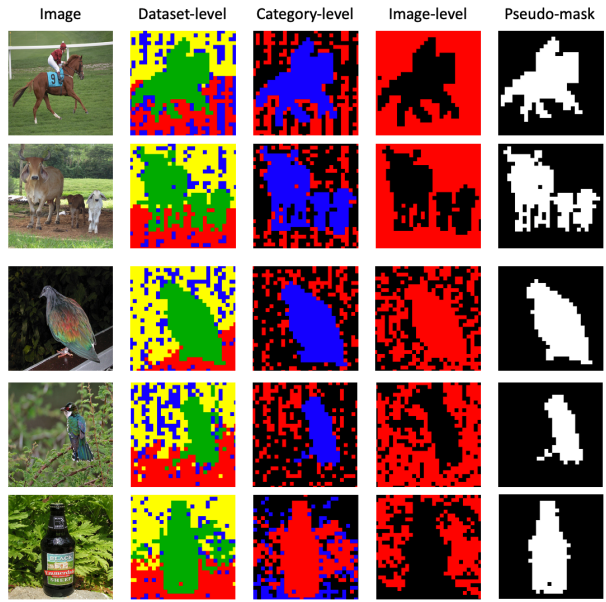}
    \caption{With simple clustering of attention features, we can obtain accurate pseudo-mask predictions. Dataset-level, Category-level, and Image-level masks are extracted by clustering features within the same dataset, superclass, and image, respectively. Masks at different levels each have their own strengths and weaknesses. In simple backgrounds, all masks deliver accurate predictions. However, segmentation becomes more challenging with complex backgrounds. Both Dataset-level and Category-level masks provide a precise estimation of object structure, whereas Image-level masks, though the coarsest, can be used to 1) identify the foreground class in Dataset-level and Category-level masks and 2) remove noise. By ensuring multilevel clustering consistency, we can obtain high-quality patch-level binary pseudo-masks that are ready for post-processing.
    }
    \label{fig1}
    \vspace{-0.2cm}
\end{figure}

Semantic segmentation is crucial for the fine-grained understanding and analysis of visual content. It has been widely applied in various domains, including medical imaging \cite{fan2020pranet, kumar2017dataset, sirinukunwattana2017gland, unet} and autonomous driving \cite{geiger2013vision,cityscapes,fu2019dual}. However, supervised methods heavily rely on large-scale labeled datasets \cite{pascal, cityscapes, mscoco, ade}. This not only incurs high labeling costs, but also limits models' effectiveness in real-world applications as they only operate on predefined categories. Weakly supervised methods \cite{ficklenet, seam, nsrom, mctformer} have been explored to reduce annotation costs, but they still require human effort. This has led to the rise of unsupervised semantic segmentation \cite{iic, picie, transfgu, stego} where segmentation is performed without relying on any ground-truth labels. 

Earlier methods for unsupervised semantic segmentation involved pixel-level self-supervised representation learning. They incorporated cross-view consistency \cite{picie, iic}, edge detection \cite{segsort, hgroup}, or saliency prior \cite{maskcontrast} in their framework. Until recently, the self-supervised Vision Transformer (ViT) has been proposed \cite{dino, dinov2}. It provides semantically rich patch-based features on unlabelled data and is widely applied in the task. Frameworks utilise features from the self-supervised ViT along with clustering \cite{acseg}, saliency model \cite{comus} and contrastive learning \cite{stego,HP} to obtain high-quality segmentation results. These methods, though with improved accuracy, are network-dependent. On one hand, substantial computational resources are still needed for model training, especially when operating on large datasets. On the other hand, extensive hyperparameter tuning are required for different datasets, which deteriorates the method's generalization capability. \vspace{0.1cm}

We propose a lightweight clustering framework for unsupervised semantic segmentation. We found that attention features of the self-supervised Vision Transformer exhibit strong foreground-background differentiability. Therefore, clustering can be used to separate foreground and background image patches. We first perform clustering at three levels: Dataset-level, by grouping features within the same dataset; Category-level, by clustering features within the same superclass; and Image-level, by grouping features within individual images. Both Dataset-level and Category-level results provide high-quality predictions. Image-level masks, despite being the coarsest, facilitate easy foreground-background detection and help with the removal of noisy regions. We thus maintain multilevel clustering consistency to extract the final binary patch-level pseudo-mask. Figure \ref{fig1}. displays the visualization results of Dataset-level, Category-level, and Image-level masks as well as the final pseudo-masks extracted. Afterwards, the patch-level pseudo-mask is upsampled to the pixel level and post-processed. To assign class labels, the image is cropped according to the object regions, and the corresponding CLS tokens are clustered into their respective classes. With the utilization of simple clustering, our network-free method not only has a low computation cost, but also achieves high-quality segmentation results. \newline

\noindent Our main contributions are summarised as: \vspace{0.2cm}
\begin{itemize}
  \item We propose a lightweight clustering framework for unsupervised semantic segmentation 
  \vspace{0.05cm}
  
  \item We ensure multilevel clustering consistency across the Dataset-level, Category-level, and Image-level to obtain high-quality pseudo-masks \vspace{0.05cm}
  
  \item Our approach achieves state-of-the-art performance on both PASCAL VOC and MS COCO datasets.  \vspace{0.05cm}
  
  \item We provide a comprehensive analysis of the self-supervised Vision Transformer features and a detailed comparison between DINO and DINOv2. 
\end{itemize}

\section{Related works}
\label{sec:formatting}

\noindent \textbf{Vison Transformer.} Transformer was first introduced in the natural language processing domain \cite{attentionallyouneed, bert, GPT}. It is exceptionally strong at capturing long-range dependencies, and its parallelizable architecture facilitates efficient training on modern hardware. Vision Transformer \cite{VIT} then brought the transformer architecture and attention mechanisms to Computer Vision \cite{segformer,segmenter, swin, maskformer}. It can capture global and local dependencies within images and is widely applied in the vision-language domain \cite{CLIP,zerotti,vilbert}. In 2021, Caron et al.\cite{dino} employed self-distillation with no labels to train the ViT. He utilised a multi-crop strategy and a momentum encoder in training the model, enabling the Vision Transformer to produce rich semantic features on unseen image data. 
The self-supervised ViT was therefore broadly adopted in unsupervised semantic segmentation. Recently, DINOv2 was proposed \cite{dinov2}. Compared to DINO, the ViT was trained on a larger dataset consisting of 142 million images. In addition, they added new regularization methods \cite{simsearch, deit} to stabilize the training process and employed a more efficient training algorithm with the use of Xformers \cite{xformers} and Pytorch2. Finally, they adopted a novel functional distillation pipeline to compress larger models into significantly smaller ones to reduce inference costs. DINOv2 currently supports two additional model types: 'large' and 'giant', which produce higher-dimensional features. The advancements in DINOv2 have significantly enhanced the feature quality of the self-supervised Vision Transformer.

\noindent \textbf{Unsupervised Semantic Segmentation.} The development of unsupervised semantic segmentation has evolved throughout the years. Early methods aim to maintain semantic consistency at the pixel level to obtain segmentation results \cite{segsort, infoseg, mutualmax, autoregressive, picie, iic}. IIC \cite{iic} maximizes the mutual information between the features of augmented images, while PiCIE \cite{picie} incorporates geometric consistency as an inductive bias to learn object representations. Other methods adopt visual prior \cite{maskcontrast, hgroup, scribblesup}, cross-view consistency \cite{selfgrouping, ctrans, fullyselfsup, looking} and continuity of video frames \cite{move} in their framework. However, performing semantic segmentation without prior knowledge is challenging. Recent works utilise the self-supervised ViT \cite{dino, dinov2} in various ways \cite{opart, maskdistill, transfgu, comus, acseg, stego, HP}. TransFGU \cite{transfgu} performs segmentation in a top-down manner by referencing class activation maps from the ViT. COMUS \cite{comus} applies saliency detectors to extract object proposals and assigns class labels to them according to image features. ACSeg \cite{acseg} adaptively maps learnable prototypes into concepts in images to obtain pixel-level segmentation. Contrastive learning has also been widely adopted alongside self-supervised ViT in the task \cite{stego,HP}. STEGO \cite{stego} utilizes knowledge distillation to learn correspondences between image features. Hyun et al.\cite{HP} has further improved the pipeline by uncovering hidden positive samples and ensuring semantic consistency within regions. Although these methods have produced a gradual improvement in segmentation results, they require substantial model training and computational resources. In contrast, our network-free framework has achieved state-of-the-art performance in unsupervised semantic segmentation, while requiring only simple clustering.
\section{Method}
\begin{figure*}[ht]
    \centering
    \includegraphics[width=\textwidth]{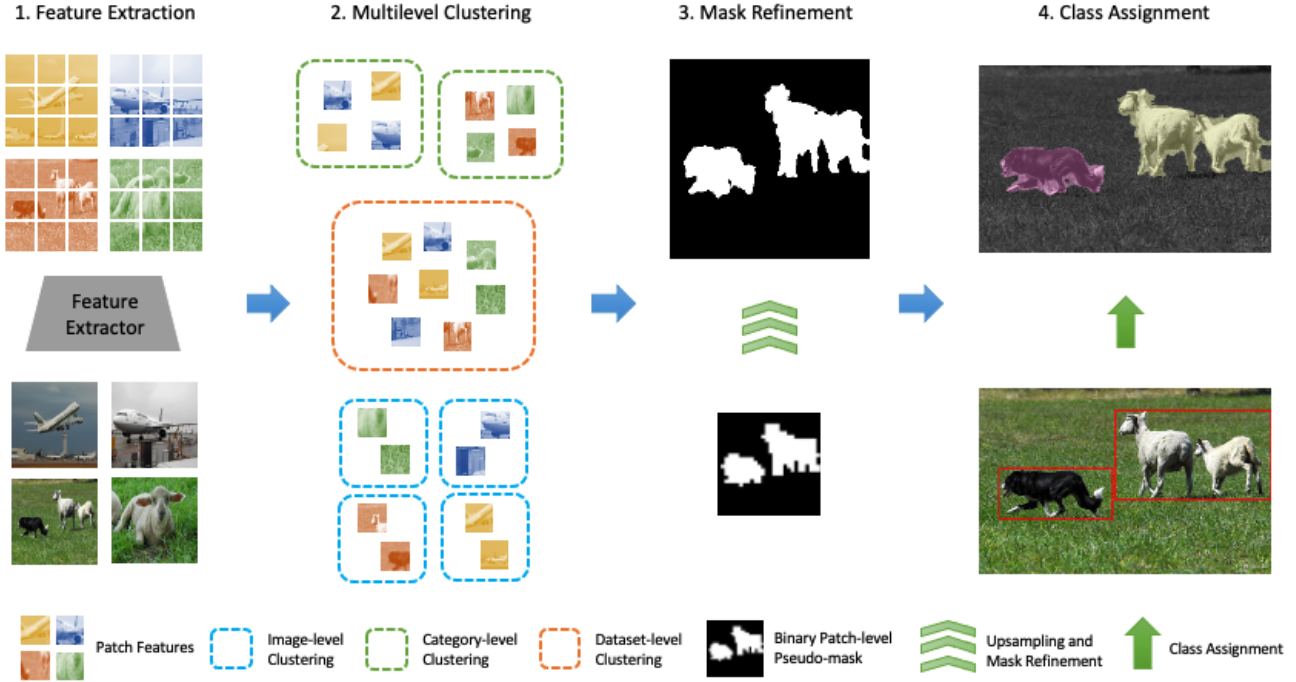}
    \caption{
    \textbf{Illustration of our Lightweight Clustering Framework.} We first utilise the self-supervised Vision Transformer to extract image patch features. Then, we perform clustering at the Image-level, Category-level, and Dataset-level. We further ensure multilevel clustering consistency and extract the binary patch-level pseudo-mask. The mask is then upsampled and refined accordingly. Finally, object regions are cropped and clustered into their respective classes.}
    \label{maindiagram}
\end{figure*}

In this section, we provide details of our lightweight clustering framework. Figure \ref{maindiagram}. illustrates the overall structure of our method.

\subsection{Preliminaries}
\label{preliminaries}
\textbf{Query-Key-Value (QKV) Attention Features.} QKV attention features in transformers \cite{attentionallyouneed, VIT} are embedded representations used for self-attention. They enable the model to query, compare, and attend to different parts of the input sequence while computing attention scores. We found that attention features of the self-supervised Vision Transformer \cite{dino,dinov2} exhibit strong foreground-background differentiability. Therefore, we can extract accurate pseudo-mask predictions by simply clustering them. In our framework, we employ cosine distance clustering on attention features because it is straightforward to implement and is suitable for handling a large number of samples. \vspace{0.1mm}

\noindent \textbf{Classification (CLS) token.} The Classification token in the Vision Transformer \cite{VIT} is added as the first token in the input sequence and provides a summarized representation of the entire image. We utilize spectral clustering to cluster CLS tokens throughout the framework since there are fewer instances, and it can effectively capture non-linear relations.

\subsection{Cosine Distance clustering}
\label{csim_clustering}
We adopt k-means clustering with cosine distance as the distance metric. Cosine similarity effectively captures similarity between vectors without being biased towards those with larger magnitudes. Cosine distance between attention feature $A$ and $B$ can be expressed as  

\begin{equation}
1 - \frac{A \cdot B}{|| A || \, || B ||}
\end{equation}

It is worth noting that k-means with cosine distance can be easily implemented with the traditional k-means. This is because cosine distance is proportional to euclidean distance when vectors are normalised.
\begin{equation}
\begin{aligned}
|| {X_{1} - X_{2}} ||^2 _2 &= X_{1}^{T}X_{1} + X_{2}^{T}X_{2} - 2X_{1}^{T}X_{2} \\
&= 2 - 2 X_{1}^{T}X_{2} \\
&= 2(1 -  X_{1}^{T}X_{2})
\end{aligned}
\end{equation}

We first apply L2 normalisation to attention features, followed by k-means clustering to perform cosine distance clustering on the feature vectors.

\subsection{Multilevel Clustering Consistency}
\label{MD_clustering}

\noindent The clustering of attention features is the main focus of our framework. We found that features clustered at different levels possess respective strengths and weaknesses, and thereby have distinct uses. Setting two, three, and four clusters for Image-level, Category-level, and Dataset-level clustering respectively, yields optimal results across all datasets.

\noindent \textbf{Image-level.} Image-level features refer to the attention features of an individual image. They are clustered into two groups: one for the foreground and the other for the background. Though binary masks produced at the image level are the coarsest, they facilitate easy identification of foreground and background regions. This assists in selecting the foreground cluster from the three and four clusters used in Category-level and Dataset-level clustering respectively; more details are provided in Sec \ref{fg_select}. In addition, Image-level masks provide a rough approximation of the object region. This allows us to remove random noises in the final pseudo-mask by excluding areas not present in the Image-level mask.

\noindent \textbf{Category-level.} 
Category-level features refer to the attention features of all samples within the same superclass. First, the CLS tokens for all images are extracted, and they're clustered into their respective superclass. PASCAL~VOC, for example, contains four superclasses: Person, Animal, Vehicle, and Indoor. Since objects in the same superclass tend to share similar backgrounds, we cluster the Category-level features in each superclass into three clusters: one for the foreground and the remaining two for the background. By correctly selecting the foreground cluster, we can obtain a precise pseudo-mask at the Category-level.

\noindent \textbf{Dataset-level.}
Dataset-level features refer to the attention features across all samples in the dataset. Since most datasets contain diverse backgrounds, we found that dividing the features into four clusters yields the best results. One cluster corresponds to the foreground, while the remaining three clusters contain background image patches. 
By accurately identifying the foreground cluster, we can obtain a high-quality pseudo-mask prediction at the Dataset-level. \vspace{0.2cm}

\noindent To ensure multilevel clustering consistency, we retain common foreground regions between Dataset-level and Category-level masks, while removing noisy areas that are not present in Image-level foreground masks. This can be formulated as: \vspace{-0.3mm}

\begin{equation}
(Dataset \cap Category) - (1 - Image)
\end{equation}

\subsection{Foreground Cluster Selection}
\label{fg_select}
 Image-level masks are used to identify the single foreground cluster in Dataset-level and Category-level clustering results. This is because they contain only two groups and enable easy differentiation between foreground and background. The foreground cluster in Dataset-level or Category-level clustering is identified based on a voting system. Confident Image-level masks with all four corners predicted as foreground or background are first  extracted. Then, a mask-flipping check is performed to ensure that the mask covers the foreground rather than the background. Finally, we select the cluster index in the Dataset-level or Category-level mask that shares the most common pixels with the binary foreground mask. This step is repeated across the entire dataset / category and the cluster index that receives the most votes is determined to be the foreground cluster. The pseudo-code is given in algorithm \ref{fgalgo}.
 
\begin{algorithm} \caption{Foreground Cluster Identification}
\label{fgalgo}
\begin{algorithmic}[1]
    \State $clusterVote = \{0:0, 1:0, 2:0, 3:0\}$
    \vspace{0.05cm}
    \For {$maskIndex \hspace{0.1cm} \textbf{in range}\hspace{0.1cm} numMasks$}
    \vspace{0.05cm}
    \State $binary = binaryMasks[maskIndex]$
    \vspace{0.05cm}
    \State $numCorners =$ \Call{checkCorner}{$binary$}
    \vspace{0.05cm}
    \If {$numCorners == \text{4}$} 
    \vspace{0.05cm}
    \State $binary =$ \Call{CheckflipMask}{$binary$}
    \vspace{0.05cm}
    \State $dataset = datasetMasks[maskIndex]$
    \vspace{0.05cm}
    \State $fgIndex = $\Call{checkColor}{$binary$, $dataset$}
    \vspace{0.05cm}
    \State $clusterVote [fgIndex] ++$
    \vspace{0.05cm}
    \EndIf
    \vspace{0.05cm}
    \EndFor
    \vspace{0.05cm}
    \State $fgCluster =$ \Call{maxKey}{$clusterVote$}
\end{algorithmic}
\end{algorithm}

\subsection{Mask refinement}
\label{Mask_refinement}
The pseudo-mask obtained from the above clustering procedure is at the patch level and of low resolution. Therefore, the mask is first upsampled to the pixel level. Small components are then removed, and Conditional Random Field (CRF) \cite{crf} is finally applied to refine the binary mask.

\subsection{Class Assignment}
\label{class_assignment}
The discovered object regions are cropped and fed into the self-supervised ViT to extract corresponding CLS tokens. These tokens are then clustered into respective classes.

\section{Experiments}

\begin{figure*}[t]
    \centering
    \includegraphics[width=1.0\textwidth]{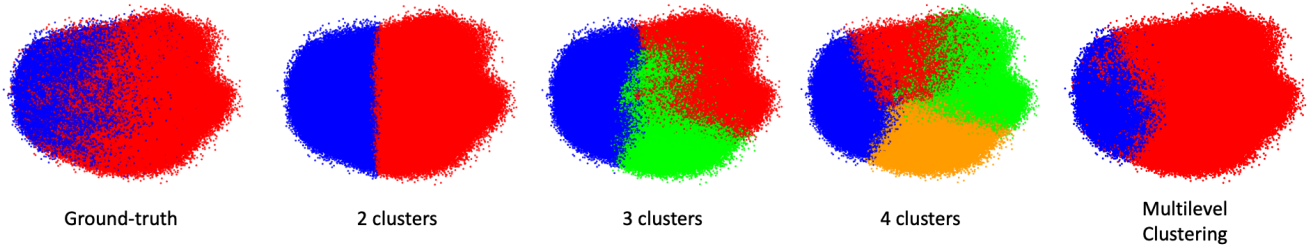}
    \caption{
    \textbf{PCA visualization of 'Key' attention features of the PASCAL VOC validation dataset.} We first showcase the Principal Component Analysis visualization of 'Key' attention features corresponding to foreground and background patches from the ground-truth PASCAL VOC validation dataset. Blue corresponds to foreground image patches, while red represents the background. Additionally, we present the results achieved by clustering the attention features into two, three, and four clusters. Finally, we display the feature distribution achieved through our multilevel clustering framework.
    }
    \label{pcavis}
    \vspace{-0.2cm}
\end{figure*}

\subsection{Datasets and Experimental Settings}
\label{expdetails}
\textbf{Datasets.} Following existing works, we first evaluated our framework on the PASCAL VOC dataset \cite{pascal}. It is a semantic segmentation dataset that contains 21 classes – 20 foreground and 1 background. The 20 object classes can be further divided into four superclasses. We also evaluated our framework on the more challenging MS COCO dataset \cite{mscoco}, which features objects in context with each other and includes 80 'thing' classes. The 80 classes can be further divided into 11 superclasses. Following the configuration of COMUS \cite{comus}, we transformed the instance segmentation masks into class masks by merging all masks of the same class, and an additional background class was added which encompass pixels not segmented.

\begin{figure*}[t]
    \centering
    \includegraphics[width=1.0\textwidth]{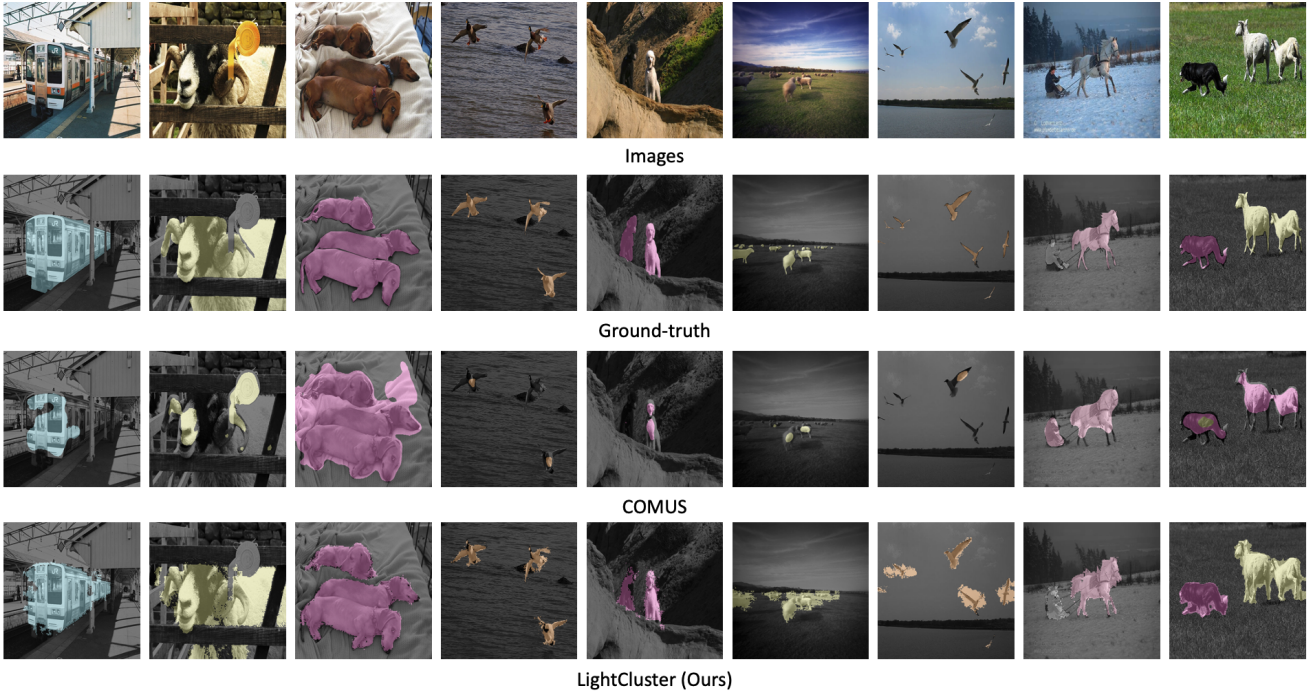}
    \caption{
    \textbf{Qualitative results on the PASCAL VOC 2012 dataset.} We display the segmentation results of our framework in comparison with the current state-of-the-art method COMUS \cite{comus} after two rounds of self-training.
    }
    \label{pascalvis}
\end{figure*}

\noindent \textbf {Evaluation Protocols and Metrics.}
The referenced benchmarks were originally employed for supervised segmentation. In the unsupervised setting, we generated a number of clusters equal to the number of classes in each dataset. Subsequently, we utilised Hungarian matching to assign classes to their respective ground-truth counterparts. To assess the quality of the segmentation results, the Mean Intersection-Over-Union (mIoU) metric was used.

\noindent \textbf {Implementation Details.} We utilised ViT-Small trained with DINOv2 \cite{dinov2} as the model to extract image features. Following existing works \cite{DSS,CutLER,maskdistill}, we utilised the 'Key' feature from the last attention layer for clustering. Previous methods resized the width and height of images to 224 and used ViT trained with DINO to extract attention features \cite{acseg,DSS,opart}. DINO ViT has a smallest patch size of 8, while DINOv2 ViT has a patch size of 14. To maintain the same feature map resolution (28x28) and fairness in comparison, we first resized images to 224x224, then upsampled them with bilinear upsampling to 392x392 to extract features. After clustering, we obtained a patch-level binary mask of 28x28, which was then upsampled with bilinear upsampling to 224x224 and post-processed. Class labels are then assigned, and the final result is evaluated at this resolution, 224x224.

\subsection{Qualitative Results}
To demonstrate the legitimacy and effectiveness of our framework, we display the visualization of 'Key' attention features of the PASCAL VOC dataset using Principal Component Analysis (PCA), as shown in Figure \ref{pcavis}. We claim that attention features exhibit strong foreground-background differentiability. From the feature distribution of the dataset, it is evident that foreground patches are mainly located on the left, whereas background patches are situated on the right. Then we present the clustering results when 'Key' attention features are grouped into two, three and four clusters. The major observations are as follows: 1) Either foreground or background patches dominate in respective clusters 2) Background patches on the right split into multiple classes, while foreground patches remain in one cluster. Furthermore, our multilevel clustering framework refines the clustering results by retaining confidence predictions and removing noisy areas. As shown in the figure, multilevel clustering could better mimic the foreground background distribution of the ground truth dataset when compared to single-level clustering. This therefore reflects multilevel clustering could bring segmentation improvements. In addition, it is important to note that while clustering attention features into two and three clusters may seem to align more closely with the ground-truth dataset, they tend to over-identify a significant number of foreground image patches. This severely deteriorates the quality of pseudo-masks obtained, as it is challenging to refine and eliminate incorrectly masked regions. 

Additionally, we present visualization results on PASCAL VOC alongside the state-of-the-art method COMUS \cite{comus} after two rounds of self-training in Figure \ref{pascalvis}. Our framework introduces four major improvements. First, it achieves perfect mask segmentation for single objects in complex backgrounds. This is due to the presence of image patches that contain similar and correlated features. Secondly, our framework has high adaptability to multi-object segmentation. It can accurately discover and segment multiple objects in an image due to its effective foreground region discovery. Moreover, our framework is exceptionally good at segmenting a large number of small objects within an image. Many existing methods adopt a network-dependent pipeline. Neural networks, however, suffer from a limited receptive field and incur lots of downsampling operations. They are thereby unable to capture contextual information of small objects. In contrast, our network-free approach bypasses these constraints and can adeptly uncover and segment small objects. Finally, our framework can produce high-quality per-image multiclass segmentation. 

\subsection{Quantitative Results}
We compare our framework against prior techniques for unsupervised segmentation, as shown in Table \ref{vocmiou}. We observe that while self-training can bring improvements in segmentation results, it requires increased computational resources and a longer training time. When evaluated on the PASCAL VOC val dataset, our method demonstrated significant improvement over the current state-of-the-art method, achieving a 4.5-point increase in mIoU, at 54.5\%. It is important to note that this comparison was made with their results after two rounds of self-training. This highlights the effectiveness and lightweight nature of our framework.

We also conducted a detailed evaluation on the MS COCO val dataset, as shown in Table \ref{mscocoresult}. Since the dataset contains 5000 images, we ran our framework at a batch size of 1000. Overall, our method achieved a higher mIoU of 21.4\%. We were also able to identify more objects, and for classes segmented with an mIoU higher than 20\%, we attained a better accuracy of 41.2\%. This demonstrates our framework has better segmentation capabilities and can uncover a broader range of object categories.

\begin{table}
    \centering
    \begin{minipage}{0.5\textwidth}
        \centering
        \renewcommand{\arraystretch}{1.1}
        \setlength{\tabcolsep}{5pt}
        \begin{tabular}{lccc}
            \hline
            \bf Method & \bf mIoU & \bf w self-training mIoU \\
            \hline
            IIC \cite{iic} & 9.8 &  / \\
            MaskContrast \cite{maskcontrast} & 35.0  & / \\
            DSM \cite{DSS} & 30.8 $ \pm $ 2.7  & 37.2 $ \pm $ 3.8\\
            Leopart \cite{opart} & 41.7  & /\\
            TransFGU \cite{transfgu} & 37.2  & /\\
            MaskDistill \cite{maskdistill}  & 42.0  & 45.8\\
            ACSeg \cite{acseg} & 47.1 $\pm$ 2.4  & /\\
            COMUS \cite{comus} &  43.8 $\pm$ 0.1 &  50.0 $\pm$ 0.4\\
            \hline
            \rowcolor{aliceblue!80} LightCluster (\emph{Ours}) & \textbf{54.5} $ \pm $ 0.3  &  /\\
            \hline
        \end{tabular}
        \caption{\textbf{Unsupervised semantic segmentation results on PASCAL VOC.} We included results with self-training.}
        \label{vocmiou}
    \end{minipage}
    \hfill
    \vspace{-11pt}
\end{table}

\begin{table*}[ht]
\label{tab:unsup_coco}
\centering
\hspace{0.24cm}
\renewcommand{\arraystretch}{1.1}
\begin{tabular}{l ccccccc}
\toprule
& all  & {} &   \multicolumn{2}{c}{discovered (with IoU$\ge 20\%$)} & {} & \multicolumn{2}{c}{have cluster (with IoU$>0\%$)} \\
\cmidrule{2-2} \cmidrule{4-5} \cmidrule{7-8} 
 & mIoU  && number &  mIoU && number &  mIoU \\
\midrule
COMUS &	18.2 && 33 & 36.6 && 73 & 20.2   \\
COMUS (w self-training) &  19.6 && 34	& 40.7  && 60 & 26.5 \\
\hline 
\rowcolor{aliceblue!80} \vspace{-0.05cm} LightCluster (\emph{Ours}) &  \textbf{21.4} && \textbf{39}	& \textbf{41.2}  && \textbf{74} & \textbf{24.4} \\
\bottomrule
\end{tabular}
\caption{\textbf{Unsupervised semantic segmentation evaluated on the MS COCO val dataset}. We present the mIoU result of our framework when evaluated on the MS COCO val dataset. We display categories with an IoU greater than 20\% from all 81 categories and regard them as 'discovered' categories. Additionally, we report the mIoU for classes with a cluster (IoU larger than zero).}
\label{mscocoresult}
\end{table*}
\section{Ablation Study}
In this section, we first provide a detailed analysis of our framework. Particularly, we explore the effectiveness of multilevel clustering, as well as the model's performance with different number of clusters. The experiments were conducted on the PASCAL VOC dataset using DINOv2 ViTS/14. Next, we conduct an ablation study to compare the performance of DINO and DINOv2, as well as examine the use of different features in the transformer. 

\subsection{Importance of Multilevel Clustering}
We provide a comprehensive evaluation of the effectiveness of multilevel clustering using the PASCAL VOC dataset. Table \ref{multiclus} presents the segmentation results when we employ Dataset-level, Category-level, and Image-level clustering progressively to extract pseudo-masks. These masks are then upsampled and refined using the same approach. From the results, we can see that utilising Dataset-level and Category-level clustering together helps yield higher-quality masks. Further improvements in mIoU and pixel accuracy are observed when noisy areas are removed with Image-level clustering.  \vspace{0.02cm} 

\begin{table}[H]
    \centering
        \centering
        \renewcommand{\arraystretch}{1.1}
        \setlength{\tabcolsep}{10pt}
        \begin{tabular}{lcc}
            \hline
            \bf Cluster Type & \bf mIoU  & \bf Pixel Accuracy\\
            \hline
            Dataset-level &  53.0 & 85.7\\
            + Category-level &  53.3 & 85.2\\
            \rowcolor{aliceblue!80} + Image-level &  54.5 & 86.7\\
            \hline
        \end{tabular}
        \caption{\textbf{Segmentation results of multilevel clustering components} We evaluate the performance of adopting different levels of clustering on the PASCAL VOC dataset.}
        \label{multiclus}
    \hfill
    \vspace{-0.7cm}
\end{table}

\begin{figure*}[ht]
    \centering
    \includegraphics[width=1.0\textwidth]{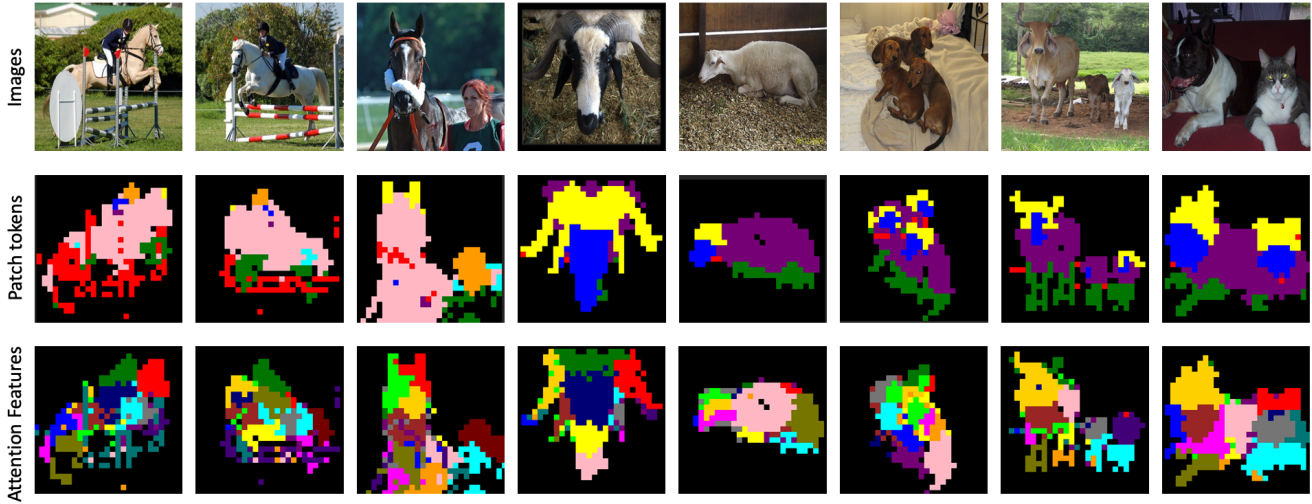}
    \caption{
    \textbf{Qualitative results using patch tokens for class assignments on the PASCAL VOC 2012 val dataset.} We present visualizations of clustering foreground patch tokens and 'Key' attention features into 20 classes. The results demonstrate that clustering based on patch tokens can accurately identify object parts, while clustering using attention features is unable to do so. The discovered object parts include human heads (orange), ears of animals (yellow), mouths of animals (blue), and legs/hands of animals (green). The bodies of horses are grouped into the pink cluster, whereas the bodies of other animals, which share a similar texture, are categorized into the purple cluster.}
    \label{fgpatch}
    \vspace{-0.0cm}
\end{figure*}

\subsection{Different cluster number}
The number of clusters is the main design choice in our framework. It plays a crucial role in ensuring foreground patches are properly grouped into one cluster, while the background shares the remaining. We display the segmentation results with different number of clusters for Dataset-level, Category-level and Image-level clustering in Table \ref{clusternum}. As shown, our framework performs optimally when four, three, and two clusters are used for Dataset-level, Category-level and Image-level clustering. \vspace{0.1cm}

\begin{table}[H]
    \centering
        \centering
        \renewcommand{\arraystretch}{1.1}
        \setlength{\tabcolsep}{4pt}
        \begin{tabular}{ccccc}
            \hline
            \bf Dataset & \bf Category & \bf Image & \bf mIoU & \bf Pixel Accuracy \\
            \hline
            3 & 3 & 2 & 53.8 & 86.3\\
            4 & 4 & 2 & 33.2 & 80.1\\
            5 & 3 & 2 & 30.9 & 79.7\\
            \rowcolor{aliceblue!80} 4 & 3 & 2 &  54.5 & 86.7\\
            \hline
        \end{tabular}
        \caption{\textbf{Segmentation result with different number of clusters} We evaluate the segmentation performance of our framework with different number of clusters on PASCAL VOC. }
        \label{clusternum}
        \vspace{-0.3cm}
    \hfill
    \vspace{-0.3cm}
\end{table}

From the table, we can also see that the performance of our framework dropped significantly when we increased the number of clusters at the Dataset-level and Category-level to 5 and 4 respectively. This is because when the attention features are divided into more clusters, foreground image patches are split among multiple clusters. Since our framework assumes that there is only one foreground cluster, many foreground patches are overlooked, which severely harms the segmentation results.

\subsection{Comparing DINO with DINOv2}
DINOv2 \cite{dinov2} was recently published by Meta. The vision transformer was trained with a larger curated dataset and a more efficient training algorithm with the help of PyTorch2 and Xformers \cite{xformers}. A newly implemented functional distillation pipeline was also introduced, which brought significant improvements to token quality. The biggest concern, however, is that DINOv2 has a patch size of 14, while DINO's smallest patch size is only 8. This might affect the resolution and quality of our pseudo-masks. In Table~\ref{dinodinov2}, we compare the performance of DINO's and DINOv2's 'Key' attention features and CLS token. We assess the effectiveness of 'Key' attention features in pseudo-mask generation and the ability of CLS tokens in class labelling. Since Category-level features initially depend on CLS tokens to perform class assignments, we use the Dataset-level mask as the final pseudo-mask to ensure accurate comparison. Upsampling and mask refinement are then followed. Object regions are finally cropped and class labels are assigned according to the corresponding CLS tokens.

\begin{table}[H]
    \centering
        \centering
        \renewcommand{\arraystretch}{1.1}
        \setlength{\tabcolsep}{5pt}
        \begin{tabular}{cccc}
            \hline
            \bf Attention feature & \bf CLS token & \bf mIoU & \bf PA \\
            \hline
            DINO ViT-S/8 & DINO ViT-S/8  &  52.8 & 84.6 \\
            DINOv2 ViT-S/14 & DINO ViT-S/8 & 51.5 & 85.4 \\
            DINO ViT-S/8 & DINOv2 ViT-S/14 &  53.4 & 84.8\\
            DINOv2 ViT-S/14 & DINOv2 ViT-S/14 & 53.0 & 85.7\\
            \hline
        \end{tabular}
        \caption{\textbf{Comparing DINO and DINOv2 performance  on PASCAL VOC.} We used the 'Key' attention features to generate binary pseudo-mask, and CLS tokens for class assignment. Experiments are conducted following configurations in Sec \ref{expdetails}.}
        \label{dinodinov2}

    \hfill
    \vspace{-15pt}
\end{table}

From the table, we can see that 1) DINO ViT can generate higher-quality pseudo-masks. This is because it has a smaller patch size, and therefore produces a higher resolution output. 2) DINOv2 ViT's CLS token provides improved class representation. It can therefore better assign object regions into respective classes. Both DINO and DINOv2 can produce masks of high pixel accuracy.

\subsection{Comparing attention features and patch tokens}
It is widely questioned why attention features, rather than patch tokens from the last layer of the ViT, are used for clustering in our framework. This is because patch tokens incorporate class-specific information, which makes them unsuitable for foreground-background clustering. Instead, if we were to first discover foreground image patches according to our clustering pipeline, then directly cluster the corresponding foreground patch tokens based on the number of targeted classes, as shown in Figure \ref{fgpatch}, object parts would be grouped correctly due to the class information patch tokens contain. On the other hand, attention features primarily capture foreground-background information. Clustering them into specific classes would result in messy outcomes. As the goal of clustering in our framework is to perform foreground-background differentiation, attention features are the better choice.

\subsection{Lightweight framework}
Our pipeline is based on simple clustering and therefore is highly efficient. On average, an experiment on PASCAL VOC takes less than 45 minutes. For existing methods that require self-training, model learning from scratch takes hours, and extensive parameter tuning is necessary.

\section{Limitation}
Although our framework achieved high segmentation performance on  PASCAL VOC and MS COCO datasets, it is important to note that both benchmarks treat background as a single class. Further explorations are needed to cluster background regions into their respective classes.
\section{Conclusion}
In this work, we introduce a lightweight clustering framework for unsupervised semantic segmentation. We found that attention features of the self-supervised Vision Transformer possess strong foreground-background differentiability. Therefore, clustering can be used to effectively separate foreground and background image patches. Dataset-level, Category-level, and Image-level masks are extracted by clustering features within the same dataset, superclass, and image respectively. Each level's results exhibit unique characteristics, and by ensuring multilevel clustering consistency, we can extract accurate binary patch-level pseudo-masks. The mask is then upsampled, refined, and finally labeled based on the clustering of object regions' CLS tokens. Our method achieves state-of-the-art results on both PASCAL VOC and MS COCO datasets. In addition, we provide a comprehensive analysis of the self-supervised ViT features and a detailed comparison between DINO and DINOv2. We hope that this work will encourage researchers in unsupervised learning to not only prioritize model accuracy but also consider the resources expended and training time. By doing so, we can make segmentation more cost-effective and efficient.
{
    \small
    \bibliographystyle{ieeenat_fullname}
    \bibliography{main}
}
\clearpage
\setcounter{page}{1}

\maketitlesupplementary
\appendix

In the supplementary material, we begin by comparing the performance of Query, Value, and Key features in Appendix \ref{qkvsec}. We then discuss different clustering techniques in Appendix \ref{clustech}. This is followed by sections on Computational Requirements in Appendix \ref{comreq}, Limitations in Appendix \ref{limitation}, and Broader Impact in Appendix \ref{broader}.

\section{Query, Value and Key attention features}
\label{qkvsec}
 Attention features of the self-supervised Vision Transformer \cite{dino, dinov2} possess strong foreground-background differentiability. Following previous works \cite{DSS,CutLER,maskdistill}, we utilise the Key feature from the last attention layer for experiments and visualization in our main paper. Here, we present a detailed performance analysis of Query, Key and Value feature vectors. 

\noindent \textbf{Quantitative Results} We evaluate the performance of our framework on PASCAL~VOC using Query, Key, and Value features. Results are shown in Table \ref{qkvpascal}.

\begin{table}[H]
    \centering
        \centering
        \renewcommand{\arraystretch}{1.1}
        \setlength{\tabcolsep}{10pt}
        \begin{tabular}{ccc}
            \hline
            \bf Attention Feature & \bf mIoU  & \bf PA \\
            \hline
            COMUS &  43.8 $\pm$ 0.1 & - \\
            COMUS (self-training) &  50.0 $\pm$ 0.4 & - \\
            LightCluster (Query) &  50.5 & 85.7\\
            LightCluster (Key) &  54.5 & 86.7\\
             \rowcolor{aliceblue!80} LightCluster (Value) &  55.7 & 87.0\\
            \hline
        \end{tabular}
        \caption{\textbf{Query, Key and Value segmentation results on PASCAL VOC} We evaluate the performance of adopting Query, Key and Value features in our framework on PASCAL VOC}
        \label{qkvpascal}
    \hfill
    \vspace{-0.7cm}
\end{table}

We can see that using Query, Key, and Value attention features in our framework can all achieve a higher mIoU than that of the current state-of-the-art method COMUS \cite{comus} after two rounds of self-training. Furthermore, a high Pixel Accuracy (PA) is achieved. This reflects our framework is sound and effective. 

\begin{figure*}[ht]
    \centering
    \includegraphics[width=1.0\textwidth]{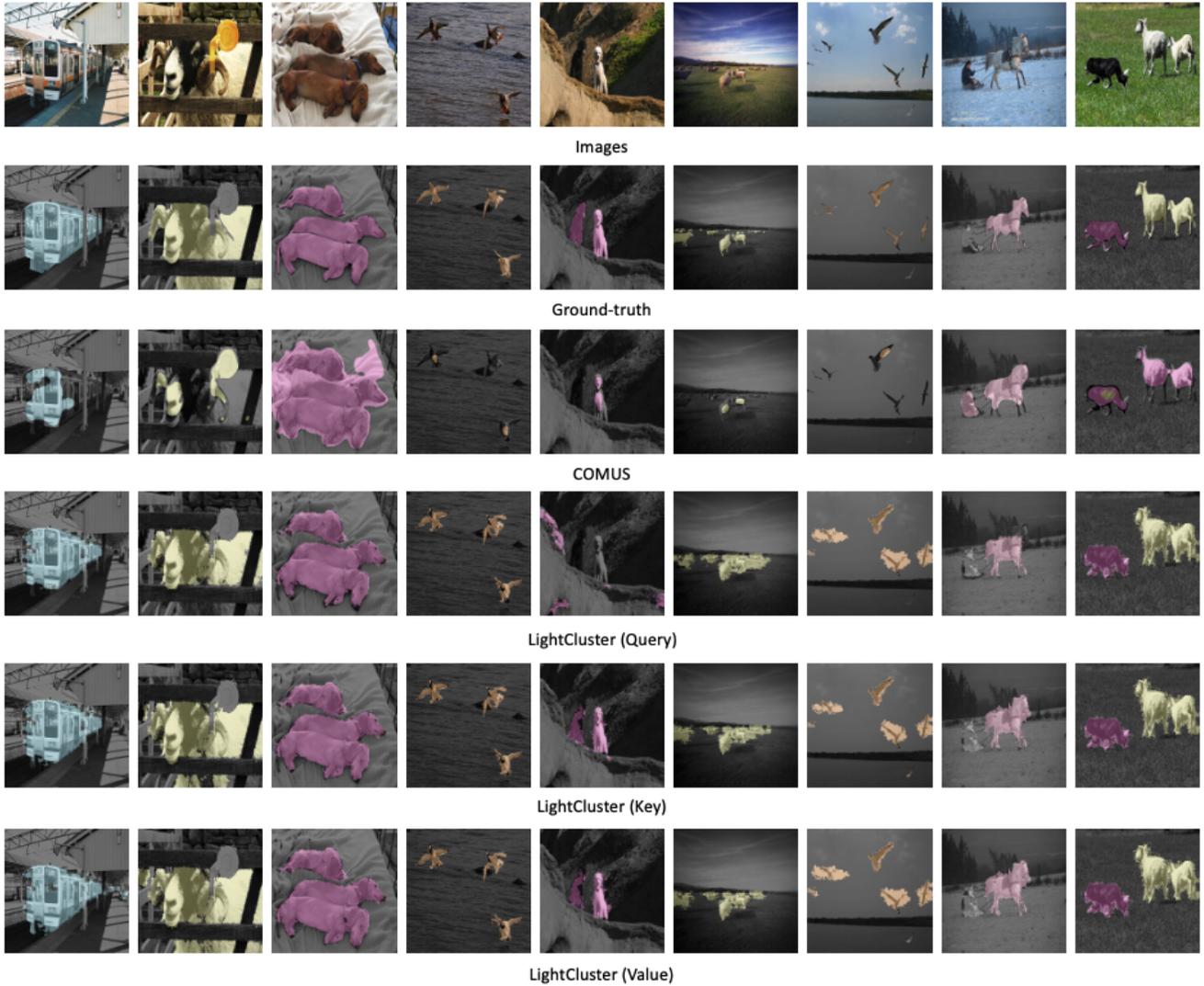}
    \caption{
    \textbf{Qualitative results of using Query, Key and Value features in our framework}. We present the qualitative results of our framework using Query, Key, and Value features in comparison with the current state-of-the-art method COMUS \cite{comus} after two rounds of self-training. The photos selected for comparison are taken from the main paper.
    }
    \label{vis1}
    \vspace{-0.0cm}
\end{figure*}

\begin{figure*}[ht!]
    \centering
    \includegraphics[width=1.0\textwidth]{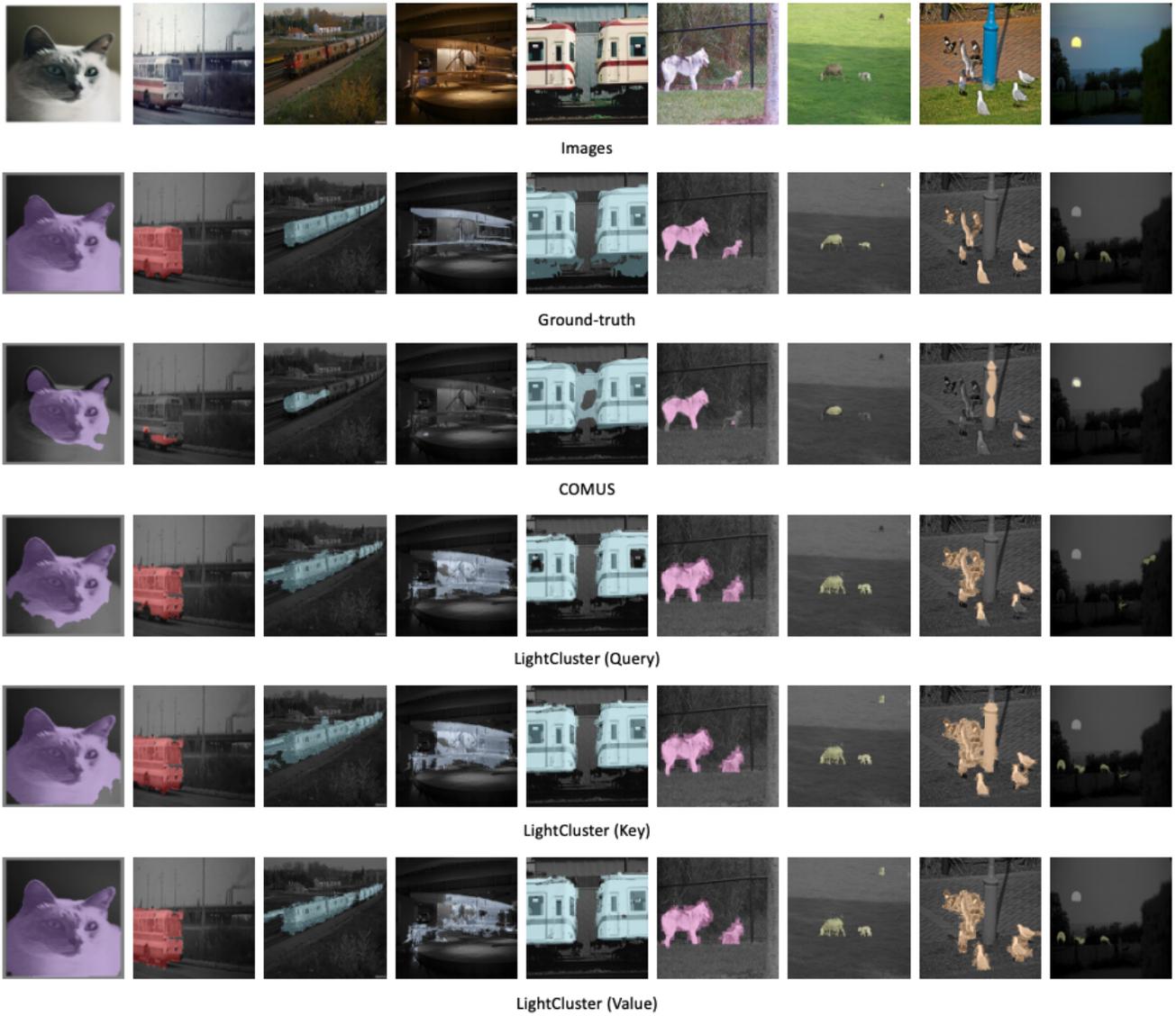}
    \caption{
    \textbf{Additional Qualitative Results on the PASCAL VOC 2012 dataset} We present additional qualitative results of our framework operating on Query, Key, and Value features, compared to the state-of-the-art method COMUS \cite{comus} after two rounds of self-training.
    }
    \label{vis2}
    \vspace{-0.0cm}
\end{figure*}

\noindent \textbf{Qualitative Results} Figures \ref{vis1} and \ref{vis2} present visualization results of incorporating Query, Key, and Value features into our framework. It is clear that they have all achieved significant improvements in segmentation results. Specifically, we observe four main enhancements:

\vspace{0.1cm}
\begin{itemize}
  \item We achieve precise mask segmentation for single objects within complex backgrounds.
  \item We demonstrate high adaptability to multi-object segmentation.
  \item We excel at segmenting a large number of small objects within a single image.
  \item We produce accurate per-image multiclass segmentation.
\end{itemize}

\noindent \textbf{Comparing between Query, Value and Key features} From the quantitative and qualitative results, we observe that Key and Value features display similar performances, and both of them outperform Query features. This is because Query features primarily guide the attention mechanism by specifying the target or context for comparison. It represents the current image patch under consideration and lacks an understanding of the broader context. As a result, masks generated from clustering are likely to be of lower quality. In contrast, the Key and Value features capture a holistic understanding of an image. Key features determine the relevance of different image patches to the current one, while Value features, weighted by these relevance scores, provide a context-aware summary. Thus, utilising Key and Value features in our framework can better differentiate between the foreground and background, and generate more accurate masks.

\noindent \textbf{PCA visualization} We display the PCA visualization results of the Query, Key, and Value feature vectors for the PASCAL VOC validation dataset, and the results obtained with our multilevel clustering framework in Figure \ref{pcasupp}. From the diagram, we observe a distinct spatial distribution of the patches. For the Query feature, the foreground patches are largely concentrated in the top right, with the remaining space primarily occupied by the background patches. For the Key and Value features, the foreground patches predominantly occupy the left side, while the background patches are mostly situated on the right. These patterns emphasize the strong foreground-background differentiability of the Query, Key, and Value features in the self-supervised Vision Transformer. Furthermore, our framework can effectively mimic the image patch distribution, thereby delivering high-quality segmentation results.

\section{Comparing different clustering techniques}
\label{clustech}
We explore the use of various techniques for clustering attention features and CLS tokens. We compare the efficacy of conventional K-means and cosine distance (CD) K-means for clustering attention features, and examine conventional K-means against spectral clustering for grouping CLS tokens on PASCAL VOC. Please note that due to the large number of attention features to be clustered, memory-intensive approaches like spectral clustering are not suitable for clustering them. The results are presented in Table \ref{diffclus}.

From the results, we can see that applying cosine distance K-means clustering to attention features and spectral clustering to CLS tokens yields the best performance.

\begin{table}[H]
    \vspace{-0.1cm}
    \centering
        \centering
        \renewcommand{\arraystretch}{1.1}
        \setlength{\tabcolsep}{10pt}
        \begin{tabular}{cccc}
            \hline
            \bf Attention & \bf CLS Token & \bf mIoU  & \bf PA\\
            \hline
            K-means &  K-means & 53.3 & 86.3 \\
            K-means &  Spectral & 53.5 & 86.4\\
            K-means (CD) &  K-means & 54.2 & 86.5\\
            \rowcolor{aliceblue!80} K-means (CD) &  Spectral & 54.5 & 86.7\\
            \hline
        \end{tabular}
        \caption{\textbf{Comparing different clustering techniques on PASCAL VOC} We evaluate our method's performance using different clustering techniques on the PASCAL VOC dataset. Experiments are conducted according to the configuration in Sec \ref{expdetails}.}
        \label{diffclus}
    \vspace{-0.5cm}
    \hfill

\end{table}
\section{Computational Requirements}
\label{comreq}
We perform all our evaluations using 2 GTX 1080 Ti GPU cards. A single run on the PASCAL VOC 2012 val dataset takes less than 45 minutes, whereas for other methods that require self-training, a complete experiment takes hours.

\section{Limitation}
\label{limitation}
The biggest limitation of our framework is class assignment. For background, it struggles to segment background regions into distinct classes. The datasets we employed, namely PASCAL~VOC \cite{pascal} and MS~COCO \cite{mscoco}, consider these areas as a single class. However, prominent semantic segmentation datasets, such as Cityscapes \cite{cityscapes}, delineate the background into multiple classes. Our framework would therefore perform worse in these scenarios. 

In terms of foreground, our framework crops out object regions and clusters the corresponding CLS tokens to perform class assignments. This method, however, cannot separate connected foreground areas into multiple classes, as only one crop is produced. Consequently, this limits our framework's performance in complex scenes.

We present failure cases in Figure \ref{failed}. The major problem with our framework stems from incorrect class assignments. More specifically, this is due to the inability of our method to correctly classify connected masks containing multiple objects into two or more distinct classes.

Nonetheless, since our framework demonstrates exceptionally high capability in extracting foreground regions with the help of attention features from the self-supervised Vision Transformer, we can still achieve state-of-the-art performance in unsupervised semantic segmentation. Further research should focus on developing better class assignment schemes for foreground and background regions.

\section{Broader Impact}
\label{broader}
One of the primary objectives of unsupervised semantic segmentation is to reduce the cost and labour used to develop segmentation networks. By eliminating the need for manually annotated pixel-level labels, unsupervised methods aim to streamline the training process. However, many existing frameworks on the task \cite{comus, maskdistill, DSS} still heavily rely on self-training to enhance segmentation performance. Training neural networks on large-scale unlabeled datasets requires substantial computational resources, thereby making segmentation equally expensive.

We hope that this work will inspire future researchers in unsupervised learning to devise novel approaches that not only achieve high segmentation accuracy but also prioritize computational efficiency and reduced resource usage. By doing so, we can effectively advance the field of unsupervised semantic segmentation, making it more accessible, cost effective, and scalable.

\clearpage
\clearpage

\clearpage
\clearpage

\clearpage
\begin{figure*}[ht]
  \begin{minipage}[b]{1.0\linewidth}
    \centering
    \includegraphics[width=1.0\textwidth]{sec/fig/pca_supp.pdf}
    \caption{
    \textbf{PCA visualization of Query, Key and Value attention features of the PASCAL VOC validation dataset.} We showcase the Principal Component Analysis visualization of the Query, Key and Value attention features corresponding to foreground and background patches from the ground-truth PASCAL VOC validation dataset. Blue corresponds to foreground image patches, while red represents the background. Then, we present the results achieved with our multilevel clustering framework.}

    \label{pcasupp}
  \end{minipage}%
    \vspace{1.0cm}

  \begin{minipage}[b]{1.0\linewidth}
    \centering
    \includegraphics[width=0.7\textwidth]{sec/fig/failed.pdf}
    \captionsetup{
      width=0.8\textwidth, 
    }
    \caption{
    \textbf{Failure samples of our framework.} The failure cases highlight the limitations of our method, specifically its inability to accurately separate connected masks into different classes.}
    \label{failed}
  \end{minipage}%
\end{figure*}
\clearpage

\newpage


\end{document}